\documentclass[sigconf,nonacm]{acmart}

\usepackage{url} 
\usepackage{amsmath,amsfonts}
\usepackage{algorithmic}
\usepackage[linesnumbered,ruled,vlined]{algorithm2e}
\usepackage{graphicx}
\usepackage{textcomp}
\usepackage{booktabs}
\usepackage[ruled,vlined]{algorithm2e}
\usepackage[table,xcdraw]{xcolor}
\usepackage{svg}
\usepackage{multirow}
\usepackage{pgfplots}
\usepackage{tikz,pgfplots}
\usepackage{subfigure}
\usepackage{enumitem}
\usepackage{xcolor}
\usepackage{pifont}
\newcommand{\fork}{\ding{55}}
\usetikzlibrary{patterns}
\pgfplotsset{compat=1.17}
\usepgfplotslibrary{groupplots}
\usepgfplotslibrary{statistics}

\definecolor{colorFst}{HTML}{bde6cd}       
\definecolor{colorSnd}{HTML}{e4eebc}       
\definecolor{colorTrd}{HTML}{fff8c5}       

\newcommand{\fs}{\cellcolor{colorFst}\bf}   
\newcommand{\nd}{\cellcolor{colorSnd}}      
\usepackage[all]{nowidow}
\def\chartheight{3.5cm}
\def\ctheight{3.7cm}
\def\BibTeX{{\rm B\kern-.05em{\sc i\kern-.025em b}\kern-.08em
    T\kern-.1667em\lower.7ex\hbox{E}\kern-.125emX}}

\newcommand{\name}{{\em AI$\mathcal{L}$FM}}

\definecolor{mymagenta}{HTML}{000000}

\AtBeginDocument{%
  \hypersetup{
    colorlinks=true,
    linkcolor=mymagenta,   
    citecolor=mymagenta,   
    urlcolor=mymagenta     
  }%
}
\setcopyright{none}

\author{%
Yixian Shen$^{1}$, 
Chaoyao Shen$^{2}$, 
Jan Deen$^{1}$, 
George Floros$^{3}$, 
Andy Pimentel$^{1}$, 
Anuj Pathania$^{1}$
}

\affiliation{%
\institution{$^{1}$University of Amsterdam, Netherlands \quad
$^{2}$Southeast University, China \quad
$^{3}$University of Thessaly, Greece}
\country{}
}

\begin{document}

\fancyhead{}                 
\renewcommand{\headrulewidth}{0pt} 
\title{Active Imitation Learning for Thermal- and Kernel-Aware LFM Inference on 3D S-NUCA Many-Cores
}

\begin{abstract}
  Large Foundation Model (LFM) inference is both memory- and compute-intensive, traditionally relying on GPUs. 
However, the limited availability and high cost have motivated the adoption of high-performance general-purpose CPUs, especially emerging 3D-stacked Static Non-Uniform Cache Architecture (3D S-NUCA) systems. 
These architectures offer enhanced bandwidth and locality but suffer from severe thermal challenges and uneven cache latencies due to 3D Networks-on-Chip (NoC). 
Optimal management of thread migration and V/f scaling is non-trivial due to LFM kernel diversity and system heterogeneity. 
Existing thermal management approaches often rely on oversimplified analytical models and lack adaptability. 
We propose~\name, an Active Imitation Learning (AIL)-based scheduling framework that learns near-optimal thermal-aware scheduling policies from Oracle demonstrations with minimal run-time overhead. 
~\name~accounts for both core-level performance heterogeneity and kernel-specific behavior in LFMs to maintain thermal safety while maximizing performance. 
Extensive experiments show that \name~outperforms state-of-the-art baselines and generalizes well across diverse LFM workloads.

\end{abstract}

\maketitle





\section{Introduction}
Transformer-based Language Foundation Models~ have achieved remarkable results across various domains, including Natural Language Processing~(NLP), Computer Vision~(CV), video processing, and multi-modal workloads. However, this high performance comes with substantial computation and data transfer costs. 
Compared to classic Convolutional Neural Networks~(CNNs), LFMs require significantly more computation-intensive operations; for instance, {\em ViT-large}~\cite{dosovitskiy2020image} demands 177x more Floating Point Operations Per Second~(FLOPs) than {\em ResNet-152}~\cite{he2016deep}.

While users commonly use Graphics Processing Units~(GPUs) and accelerators for transformer training and inference, the industry is increasingly focusing on general-purpose Central Processing Units~(CPUs) for Deep Neural Network~(DNN) inference due to their low cost and high availability in data centers. 
Furthermore, CPUs are more flexible~(programmable) and have a far more robust hardware/software ecosystem than GPUs and accelerators~\cite{hazelwood2018applied}.

Recent advancements, such as the {\em Intel Xeon CPU Max} series~\cite{kuper2024quantitative}, demonstrate the potential of x86 CPUs in accelerating Machine Learning~(ML) workloads. 
Consequently, administrators now widely deploy multi-/many-core processors for DNN inference across servers, edge devices, and commercial Systems-on-Chips~(SoCs)~\cite{liu2019optimizing}, with solutions like the {\em NXP Layerscape LX2160} and {\em ARM Cortex-A72 CPUs}~\cite{perryman2023evaluation}. 
{\em Meta}, for example, demonstrates CPU-based inference for its LLaMA~3 models~\cite{intel_llama3_cpu,wu2019machine}, leveraging CPUs’ ubiquity across data centers and edge devices, while GPUs are primarily reserved for training.
Therefore, CPUs offer broad optimization opportunities, enhancing applicability across several ML tasks.

Limited off-chip memory bandwidth and throughput are the primary bottlenecks in executing low-latency LFM inference on CPUs. 
LFMs are memory-intensive, high-throughput workloads requiring significant data movement between the CPU and memory during forward and backward passes. 
These models exhibit large memory footprints due to dynamically generated intermediate data in each layer, particularly within multi-head self-attention and feed-forward blocks. 
Stacks of encoder and decoder layers produce substantial data, such as attention scores, key, query, and value matrices, that must be stored and transferred to subsequent layers. 
This intensive data generation quickly saturates the memory bandwidth of conventional Dynamic Random-Access Memory~(DRAM), causing frequent CPU stalls as cores wait for data, ultimately degrading system performance and increasing inference time.

\begin{figure}
    \centering
    \includegraphics[width=0.97\linewidth]{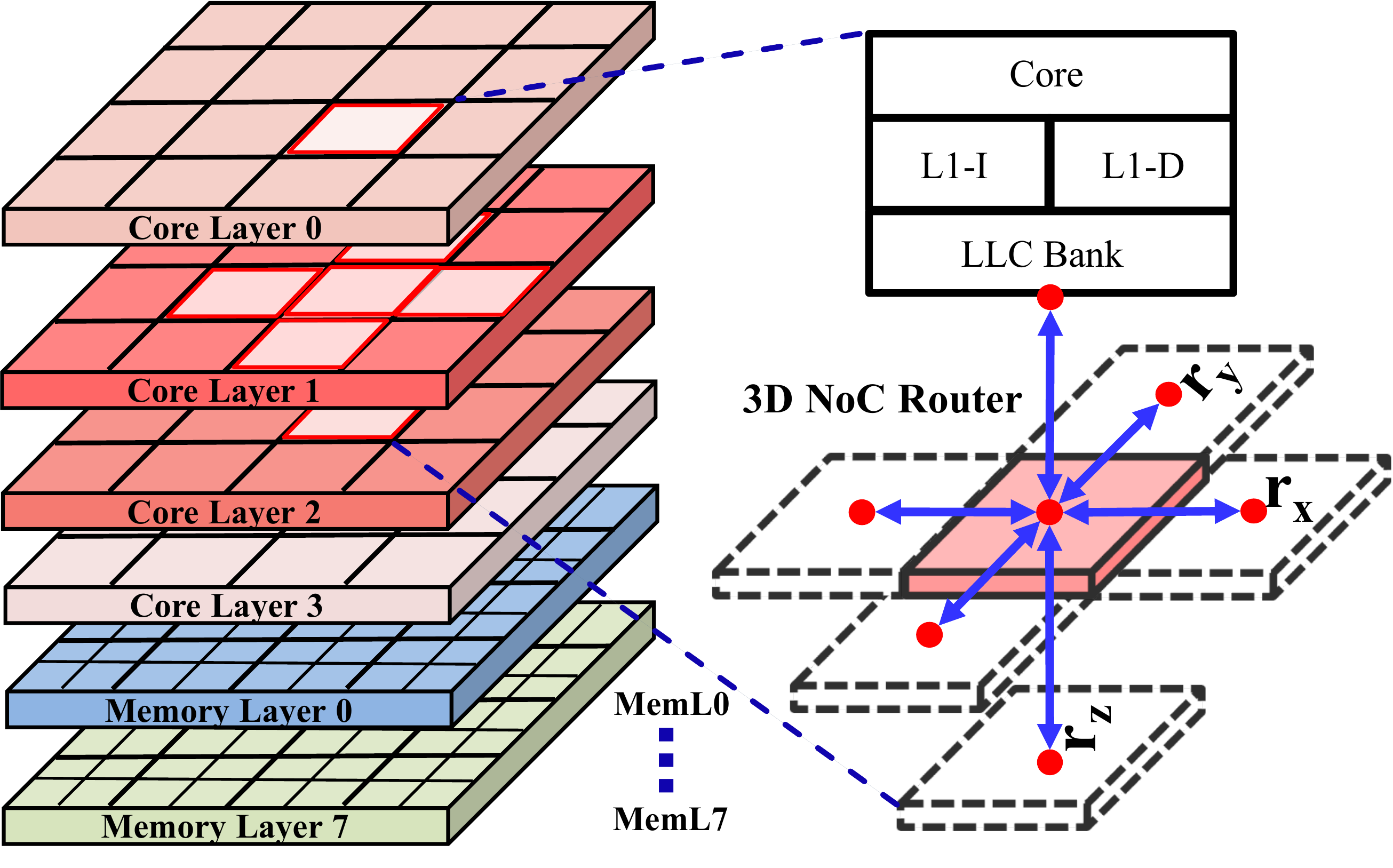}
    \caption{\small An abstract representation of 3D-stacked S-NUCA systems.}
    \label{fig:snuca_architecture}
    \vspace {-0.5cm}
\end{figure}

3D-stacked Static Non-Uniform Cache Architecture (3D S-NUCA) processor-memory systems integrate High-Bandwidth Memory (HBM) atop S-NUCA multi-/many-core processors, enhancing data locality, reducing latency, and minimizing cache coherence overhead~\cite{shen2022tcps} via logically-shared, physically-distributed Last-Level Cache (LLC). 
S-NUCA employs a static LLC to DRAM memory mapping for cache coherency~\cite{huh2005nuca,merino2008sp}. 
Figure~\ref{fig:snuca_architecture} shows an example 3D S-NUCA system equipped with a 3D Network-on-Chip~(NoC) for low-latency, low-congestion on-chip intra-processor cache coherency traffic. 
However, stacking silicon layers in 3D-stacked systems significantly increases volume without expanding surface area~\cite{siddhu2022comet}, leading to higher power density but limited heat dissipation~\cite{hsieh2013thermal,meng2012optimizing}. 
Their high bandwidth efficiency drives intensive core and memory activity, further escalating heat generation compared to non-stacked planar systems~\cite{lo2016thermal}. 
Consequently, thermal challenges in 3D S-NUCA systems often necessitate frequency reductions using Dynamic Voltage and Frequency Scaling (DVFS), which limits their performance.

\begin{figure}
    \centering
    \includegraphics[width=\linewidth]{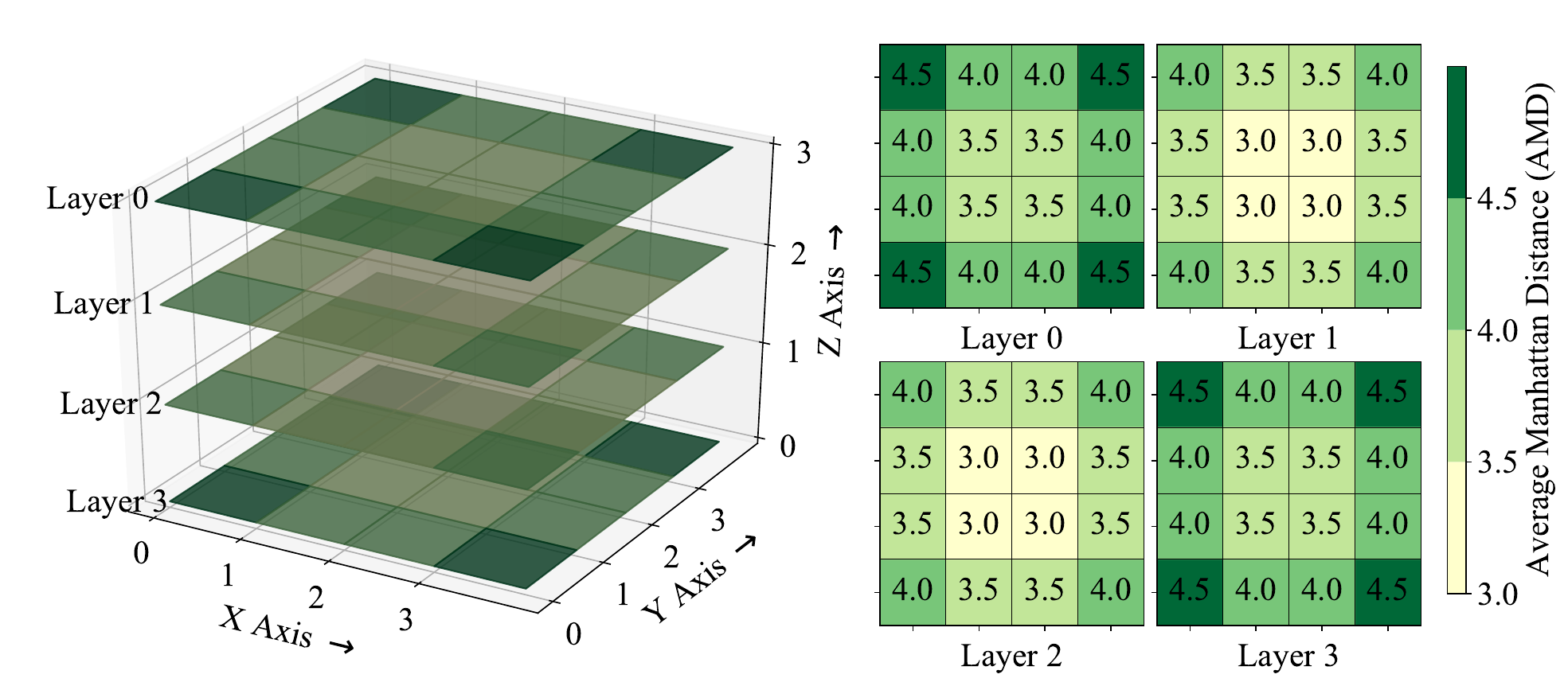}
    \small
    \caption{\small Processing core AMDs in 3D-stacked S-NUCA processors}
    \label{fig:amd}
    \vspace{-0.5cm}
\end{figure}

Furthermore, micro-architecturally identical cores in processors with 3D S-NUCA caches exhibit noticeable performance heterogeneity due to the non-uniformity in the LLC access latency, as depicted in Figure~\ref{fig:amd}. 
The Average Manhattan Distance~(AMD)~\cite{pathania2018task} for the cores is the standard proxy for measuring the core performance heterogeneity in S-NUCA processors, quantified as the average hop count of a given core from every other core. 
Cores near the center of the processor have lower AMD than those in the corners. Cores with lower AMD perform better due to a lower average LLC access latency than higher AMD cores. 
However, cores also become more thermally constrained as they get spatially closer to the processor center due to higher intra-processor heat convection from other cores and lower external convection to the ambient. 

We propose an \textbf{Active Imitation Learning (AIL)-based scheduling framework} for thermal management in 3D S-NUCA systems called \name~(AIL for LFMs). \name~characterizes the diverse computational and memory access patterns of LFM kernels and maps them to heterogeneous processing cores of 3D S-NUCA systems. 
While researchers have applied Imitation Learning (IL) to power optimization for traditional parallel applications on 2D processors~\cite{mandal2019dynamic,kim2017imitation}, extending it to thermal optimization in 3D S-NUCA systems for LFM inference is significantly more challenging due to spatial~(heat transfer), temporal (heat capacity) and workload-level (kernel diversity in LFMs) complexities. 
By integrating active learning in IL, \name~selects representative examples efficiently, reducing the sampling effort required for learning from an Oracle policy while enabling autonomous decision-making. 
Experiments validate \name's superior performance over state-of-the-art baselines and its effective generalization to diverse DNN workloads.

\textbf{Our Novel Contributions:} 

\begin{itemize}[leftmargin=2.2em, topsep=0pt, itemsep=0pt]
    \item We systematically exploit the performance heterogeneity in 3D S-NUCA systems for LFMs. By characterizing the distinct computational and memory demands of LFM kernels, we optimize their placement and execution across heterogeneous cores of 3D S-NUCA systems.

    
    \item We design, train, and employ an AIL-based framework, \name, that ensures thermal safety while maximizing the inference performance of LFMs on 3D S-NUCA systems. \name~follows Oracle policies, a trusted source, while performing autonomous decision-making, which enables near-optimal scheduling with minimal run-time overhead.



    \item We empirically validate the efficacy of \name~through extensive experiments, showing that it outperforms state-of-the-art baselines in both thermal efficiency and inference performance.  Furthermore, we demonstrate that the proposed framework generalizes effectively to unseen LFM workloads, showing robustness and adaptability.
\end{itemize}

\section{Related Work}

Recent studies have increasingly explored Large Foundation Model (LFM) inference on general-purpose CPUs as a practical complement to GPUs, motivated by limited GPU availability, deployment cost, and the desire to better utilize datacenter CPU resources.
In this direction, prior work has proposed a range of optimizations for efficient CPU deployment of LFMs, including large language models (LLMs)~\cite{tao2024robustness,huang2024optimizing,huang2025gradient,shen2025macp,shen2025ssh,zhu2025interactive}, multimodal models~\cite{huang2025image2text2image,huang2024novel,shen2025altgen,bi2025adadcp}, and reasoning models~\cite{wang2025reasoning,shen2022tcps,shenefficient}, as discussed in~\cite{shen2023efficient,zhang2025neuroada}.
He \textit{et al.}~\cite{he2024inference}, for example, improve LFM inference efficiency by reducing the size of the key/value (K/V) matrices and refining model parameters.
However, these studies primarily target conventional 2D processor platforms and focus on inference acceleration, memory reduction, or parameter efficiency, rather than the thermal and cache-locality challenges introduced by 3D-stacked architectures.
Consequently, they do not address the joint problem of temperature management, non-uniform last-level-cache access, and kernel-dependent scheduling during LFM inference on 3D S-NUCA many-core systems.

In parallel, thermal management for 3D-stacked systems has become increasingly important with the broader adoption of 3D integration technologies~\cite{kumar2017fighting,siddhu2020leakage,gourdoumanis2026multi}.
Prior work has explored several major control knobs for mitigating thermal challenges, including thermal-aware task mapping~\cite{liu2022tb,tsai2012thermal,chaturvedi2014thermal,wasala2025energy}, voltage/frequency (V/f) allocation~\cite{henkel2018dynamic,aghapour2024piqi}, and thread migration~\cite{sikal2023machine,shen2023thermal}.
Several of these studies further employ machine learning techniques for thermal management~\cite{mandal2019dynamic,kim2017imitation,zhang2017machine}, but they are generally designed for traditional parallel applications or generic many-core workloads rather than modern, resource-intensive LFMs.
Pandey \textit{et al.}~\cite{pandey2024neurotap} propose thermal-aware task mapping for DNN workloads on 2.5D processor--memory systems, which is closely related to our setting.
However, their work does not consider the kernel diversity of LFMs, the cache-distance heterogeneity of 3D S-NUCA systems, or the coupled scheduling trade-off between thermal safety and kernel-sensitive performance.
Therefore, to the best of our knowledge, prior work has not explicitly addressed thermal-aware scheduling for LFM inference on 3D S-NUCA many-core systems by jointly considering kernel-level execution behavior, cache-distance heterogeneity, and thermal constraints.

\section{Problem Formulation}
The target system is a 3D-stacked S-NUCA processor-memory system consisting of $N$ cores and $M$ memory banks. 
Each core is homogeneous and accesses a logically shared but physically distributed LLC. 
The power consumption of each core $i$, denoted as $p_i$, includes dynamic and leakage components. The steady-state temperature of the cores, represented as $T = [T_i]_{N \times 1}$, is estimated using the RC-thermal model~\cite{huang2006hotspot}.

Each target application, an LFM, is structured into computational units referred to as \textit{kernels}, each performing distinct computational tasks. Specifically, an LFM instance decomposes into an embedding kernel, $N_b$ homogeneous attention and Feed-Forward Network (FFN) kernels, and a Language Modeling (LM) head kernel.
Consequently, an application can be represented as a sequence of kernels: $\mathbf{K} = [{\mathbf{K}_i}]_{\kappa \times 1}$, where $\kappa$ is the total number of kernels in an LFM.



The optimization objective is to minimize the LFM inference time while ensuring thermal safety.
Thermal safety requires the peak core temperature $T_{\text{peak}}$ to remain below a predefined threshold $T_{\text{th}}$. 
Thread migration and transient power budgeting, where 3D-TTP~\cite{niknam20233d} conducts DVFS to mitigate thermal emergencies, are available on-chip thermal management knobs.
However, these mechanisms come with performance penalties: thread migration leads to cold-start cache overhead due to the distributed LLC, while DVFS proactively reduces throughput by lowering the core frequency.
We formalize our optimization objective as follows:

\begin{equation}
\label{eq:thermal-objective}
\begin{aligned}
&\min_{\pi_\theta} \quad & M(\pi_\theta) + \min\left(\mathcal{O}_{\text{mig}}(\pi_\theta), \mathcal{O}_{\text{dvfs}}(\pi_\theta)\right) \\
&\text{subject to} \quad & T_{\text{peak}}(\pi_\theta) \leq T_{\text{th}}
\end{aligned}
\end{equation}

where $M(\pi_\theta)$ is the total inference execution time under scheduling policy $\pi_\theta$, and $\mathcal{O}_{\text{mig}}(\pi_\theta)$, $\mathcal{O}_{\text{dvfs}}(\pi_\theta)$ represent the overheads of thread migration and DVFS, respectively. 
\name~distributes these kernels across the cores, with their performance and thermal behavior influenced by resource allocation and migration decisions.

\section{LFM Kernel Characterization}
\label{sec:kernel-profiling}
We experimentally analyze the sensitivity of LFM kernel performance to the AMD of their assigned cores. Specifically, we measure Instructions Per Second (IPS) and LLC Misses Per Kilo Instructions (MPKI) for four key LFM kernels, Embedding, Attention, FFN, and LM Head, using {\em ViT-base} workloads executed at a fixed frequency of $f = 3\,\text{GHz}$. The AMD values range from 3.0 to 4.5, as exhibited in Figure~\ref{fig:amd}.

\begin{table}[t]
\centering
\caption{\small IPS (\texttimes 10\textsuperscript{9}) and LLC MPKI of ViT kernels at varying AMDs.}
\vspace{-10pt}
\label{tab:ips_vs_amd}
\begin{tabular}{lcccccccc}
\toprule
\multirow{2}{*}{\textbf{Kernel}} & \multicolumn{4}{c}{\textbf{IPS$\uparrow$}} & \multicolumn{4}{c}{\textbf{LLC MPKI$\downarrow$}} \\
\cmidrule(lr){2-5} \cmidrule(lr){6-9}
& 3.0 & 3.5 & 4.0 & 4.5 & 3.0 & 3.5 & 4.0 & 4.5 \\
\midrule
Embedding   & 6.23 & 6.01 & 5.82 & 5.51 & 10 & 13 & 17 & 22 \\
Self-Attention   & \fs7.92 & \fs6.83 & \fs6.03 & \fs5.10 & \fs21 & \fs25 & \fs32 & \fs46 \\
FFN         & 6.59 & 6.42 & 6.21 & 6.14 & 7  & 7  & 9  & 11 \\
LM Head     &\nd5.77 & \nd5.29 & \nd4.94 & \nd4.36 & \nd11 & \nd15 & \nd19 & \nd37 \\
\bottomrule
\end{tabular}
\end{table}

Table~\ref{tab:ips_vs_amd} reveals distinct kernel sensitivity patterns to AMD. The Self-Attention kernel demonstrates the highest sensitivity, experiencing an IPS reduction exceeding 35\% as AMD rises from 3.0 to 4.5. This degradation arises from the kernel's reliance on frequent K/V cache accesses and temporal data reuse. Conversely, the FFN kernel, characterized by intensive compute operations and predominantly sequential memory accesses, exhibits minimal sensitivity, with less than a 6.5\% IPS decrease across the same AMD span. The Embedding and LM Head kernels show moderate sensitivity due to cache footprints and partial reuse characteristics, resulting in intermediate IPS degradations of  12\% to 24\%, respectively.

Beyond {\em ViT-base}, we observe consistent performance sensitivity patterns across a variety of auto-regressive and bidirectional LFMs, including {\em BERT-base}, {\em LLaMA3.2-1B}, {\em Gemma-2B}, and  {\em DeepSeek-2.4B}, which we omit for space. Notably, attention kernels are highly sensitive to cache locality, consistently exceeding 30\% IPS degradation at higher AMD values. These results underline the universality of cache-sensitive behavior across diverse LFMs, validating the general applicability of our kernel-specific mapping strategy.



\section{LFM Kernel and thermal-aware scheduling}

\begin{figure*}
    \centering
    \includegraphics[width=0.97\linewidth]{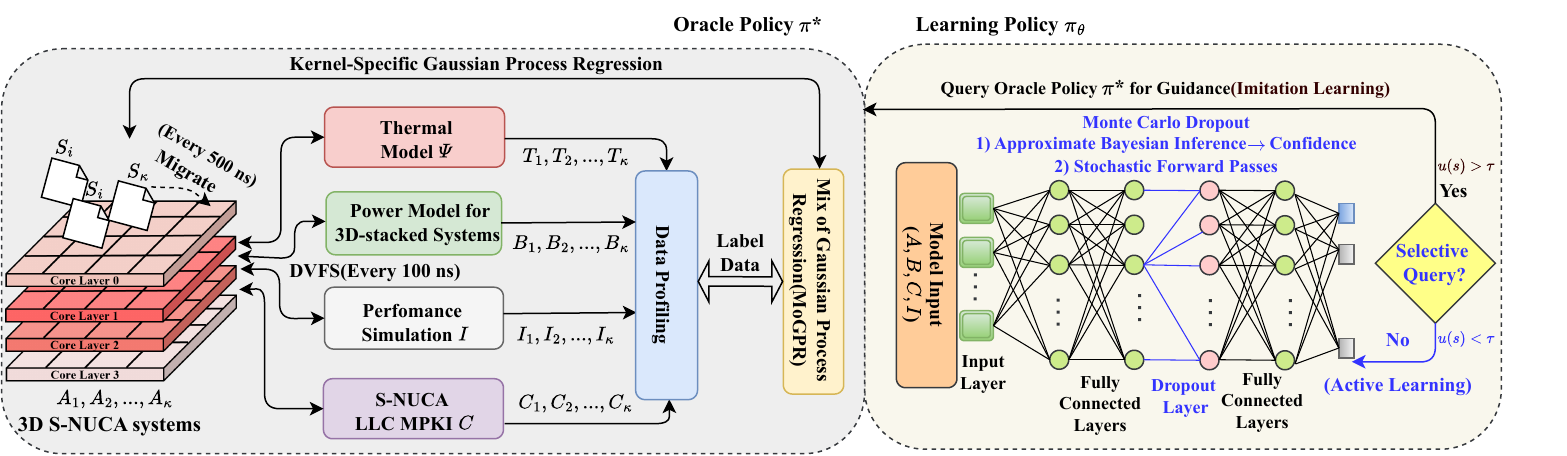}   
    \vspace{-10pt}
    \caption{\small Overview of the Oracle and Learning Policies in \name~for thermal management of 3D S-NUCA systems.
The Oracle Policy $\pi^{*}$ uses MoGPR to model MPKI--IPS--thermal dynamics and label kernel-migration utilities.
The Learning Policy $\pi_\theta$ applies MC Dropout for uncertainty estimation, enabling selective Oracle queries and confident autonomous actions.
    }
    \label{fig:tilpm}
\end{figure*}

The spatial heterogeneity in cache access latency inherent to 3D S-NUCA systems, coupled with the diverse sensitivity of LFM kernels to LLC performance, introduces significant challenges for dynamic thermal management. We design \name~(depicted in Figure~\ref{fig:tilpm}), an AIL-driven dynamic scheduling framework that holistically incorporates thermal awareness and kernel-level execution behavior. 


\subsection{Feature Selection}

Effective feature selection is critical for thermal-safe, high-performance thread migration. 
We considered IPS, MPKI, AMD, power budget, temperature, frequency, and migration history; the latter two were redundant or added overhead without improving prediction accuracy, so only the first five are retained as observable system-state features. Specifically, 
Table~\ref{tab:ips_vs_amd} shows that different LLM kernels exhibit varying sensitivity to AMD, motivating the inclusion of AMD values $\mathbf{A} = {[\mathbf{A}_i]_{\kappa \times 1}}$ as a feature for each active thread.

We include IPS $\mathbf{I} = {[\mathbf{I}_i]_{\kappa \times 1}}$ as a direct indicator of computational throughput, serving as the primary performance metric. We incorporate S-NUCA’s LLC MPKI, denoted by $\mathbf{C} = {[\mathbf{C}_i]_{\kappa \times 1}}$, to capture cache locality behavior. Higher MPKI values indicate lower spatial or temporal locality and increased dependence on external memory, amplifying the performance implications of cache placement and migration in the 3D S-NUCA systems.

Additionally, we use the active thread’s power budget $\mathbf{B} = {[\mathbf{B}_i]_{\kappa \times 1}}$, computed using the {\em 3D-TTP} power budgeting framework~\cite{niknam20233d}, which captures thermally constrained power allocation. This budgeting facilitates V/f adjustments using DVFS. Lastly, the threshold temperature $T_{th}$ and current peak temperature $T_{peak}$ define thermal operating limits, ensuring the system remains within safe bounds.

\subsection{Training Data Collection}
We collect execution traces of LFMs under diverse operating conditions, varying in both cache locality and power constraints. Concretely, for each of LLM kernel types, we define 40 distinct operating points by combining four AMD levels (3.0, 3.5, 4.0, 4.5) with ten power budgets uniformly distributed between 1.0\,W and 5.5\,W. At each operating point, we record performance traces, including IPS, MPKI, power budget, and AMD. We generate traces from {\em ViT-base}, {\em BERT-base}, and {\em LLaMA3.2-1B} under a thermal threshold of $T_{\text{th}} = 70$~\textcelsius~ and input sequence length $L = 256$.

Since execution duration may vary across operating points, we normalize all traces by dividing each into 200 equally sized temporal slices. Each slice captures a fixed execution interval characterized by a specific kernel type, AMD level, and power budget. We pair each slice with all other slices of the same kernel type but from different operating points, modeling hypothetical transitions induced by migration or DVFS actions. For each kernel, this pairing yields up to \(40 \times 39 = 1{,}560\) training examples per LLM. Across the three models, we obtain \(4{,}680\) training samples per kernel type.

 \subsection{MoGPR-Based Oracle Demonstrations}
Aggressive power budgeting reduces core activity and degrades performance; migrating threads to cooler cores can sustain throughput, but in 3D S-NUCA systems with distributed LLCs, such moves incur cold-start penalties due to lost cache locality.
Critically, migration decisions are highly kernel-dependent due to the diverse sensitivities of LLM kernels to cache distance and thermal dynamics (Table~\ref{tab:ips_vs_amd}).
Moreover, LFMs execute using model parallelism on 3D S-NUCA systems, where kernels operate in a fixed, sequential order. This structured execution enables deterministic identification of the currently active kernel at each scheduling epoch. Leveraging this property, we design kernel-specific Oracle demonstrations using a MoGPR (Mixture of Gaussian Process Regression).

\begin{figure}
    \centering
    \includegraphics[width=0.99\linewidth]{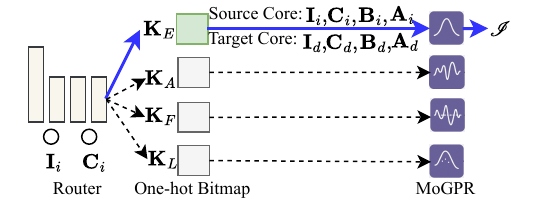}
    \vspace{-10pt}
    \caption{ \small
Illustration of the MoGPR-based  Oracle in \name. 
The selected $\mathcal{GP}$ predicts migration utility from source and target core.}
    \label{fig:mog}
\end{figure}

Figure~\ref{fig:mog} provides a concrete example to illustrate the MoGPR Oracle. We implement a lightweight yet effective routing mechanism to identify the currently executing kernel type based on run-time profiling signals. Specifically, we use a weighted combination of IPS and MPKI, $\lambda_1 \cdot \mathbf{I}_i + \lambda_2 \cdot \mathbf{C}_i$, where the coefficients $\lambda_1$ and $\lambda_2$ are learned from data to reflect the performance and memory access characteristics of each LLM kernel. Once the current kernel $\mathbf{K}_t$ is identified (e.g., $\mathbf{K}_E$ for an Embedding kernel), the corresponding GPR expert model $\mathcal{GP}_{k_t}$ is selected via a one-hot encoded bitmap gating function:

\begin{small}

\begin{equation}
\mathcal{R}(\mathbf{I_k,C_k}) =
\begin{cases}
1, & \text{if } k = \mathbf{K}_t \\
0, & \text{otherwise}
\end{cases}
\quad \text{such that } \sum_k \mathcal{R}(\mathbf{I_k,C_k}) = 1
\end{equation}
\end{small}

The utility of a migration decision in 3D S-NUCA systems depends on multiple system-level factors, including potential performance gain, cache access locality, and power budget. We define the input to each kernel-specific expert $\mathcal{GP}_k$ as a pair of run-time state vectors representing the source and destination cores. These vectors encode the observed and expected system behavior before and after migration. We model the migration utility $\mathcal{I}$ as follows:

\begin{small}
\begin{equation}
\label{eq:utility-gpr}
\mathcal{I} \sim \mathcal{GP}\left( m_k(\mathbf{x_d}), \, k_k(\mathbf{x_i}, \mathbf{x_d}) \right)
\end{equation} 
\end{small}

where $m_k(\cdot)$ denotes the mean function and $k_k(\cdot, \cdot)$ represents the covariance kernel for kernel type $k$, which may use a squared exponential kernel~\cite{seeger2004gaussian}. The input vector $\mathbf{x_i} = \left[ \mathbf{I}_i, \mathbf{C}_i,  \mathbf{A}_i, \mathbf{B}_i \right]$ represents the run-time state of the source core. Similarly, the vector $\mathbf{x_d} = \left[ \mathbf{I}_d, \mathbf{C}_d,  \mathbf{A}_d,  \mathbf{B}_d \right]$ represents the target core. Once the expert model $\mathcal{GP}_k$ predicts the expected migration utility $\mathcal{I}$, Oracle evaluates its viability, considering migration a viable candidate if the expected performance gain exceeds the estimated overhead, i.e., when $\mathcal{I} > 0$. 
 The Oracle identifies the best possible migration among all candidates by evaluating their thermal safety and potential $\mathcal{I}$ improvement at each iteration.
If multiple candidates satisfy the requirements, the Oracle selects the one that yields the highest $\mathcal{I}$ gain by the equation below.

\begin{small}
\begin{equation}
\label{eq:migrate}
\mathcal{I} = \arg\max_{\mathcal{I}_d} \left(\mathcal{I} \;\land\; T_{\text{peak}} < T_{\text{th}}\right)
\end{equation}   
\end{small}

\begin{figure*}[!t]
\begin{tikzpicture}
\footnotesize
\begin{groupplot}[
    group style={
        group size=5 by 1, 
        vertical sep=1.5cm, 
    },
    width=4.1cm,
    height=3.7cm,
    ylabel style={align=center},
    xlabel style={yshift=-5pt},
    every axis plot/.append style={line width=1pt},
]

\nextgroupplot[
    ylabel=Norm. Exec.,
    xtick={0.25, 0.5, 0.75, 1.0},
    xmin=0.2, xmax=1.05,
    xticklabels={128, 256, 512, 1024},
    xticklabel style={rotate=25, anchor=east, font=\scriptsize, xshift=0.4em,
    yshift=-0.4em},
    ymin=0.95, ymax=2.1
]

\addplot[orange, mark=*, mark size=1.5pt, dash pattern=on 3pt off 5pt, line width=0.4pt] coordinates {
    (0.25,1.00) (0.5,1.10) (0.75,1.22) (1.0,1.35)
};

\addplot[cyan, mark=triangle*, mark size=1.5pt, dash pattern=on 3pt off 5pt, line width=0.4pt] coordinates {
    (0.25,1.05) (0.5,1.15) (0.75,1.30) (1.0,1.50)
    
};

\addplot[blue, mark=square*, mark size=1.3pt, dash pattern=on 3pt off 5pt, line width=0.4pt] coordinates {
    (0.25,1.15) (0.5,1.33) (0.75,1.52) (1.0,1.85)
};

\addplot[gray, mark=pentagon*, mark size=1.5pt, dash pattern=on 3pt off 5pt, line width=0.4pt] coordinates {
    (0.25,1.25) (0.5,1.55) (0.75,1.73) (1.0,2.05)
};

\addplot[pink, mark=diamond*, mark size=1.5pt, dash pattern=on 3pt off 5pt, line width=0.4pt] coordinates {
    (0.25,1.10) (0.5,1.25) (0.75,1.45) (1.0,1.70)
};

\nextgroupplot[
    ylabel=Norm. Exec.,
    xtick={0.25, 0.5, 0.75, 1.0},
    xmin=0.2, xmax=1.05,
    xticklabels={128, 256, 512, 1024},
    xticklabel style={rotate=25, anchor=east, font=\scriptsize, xshift=0.4em,
    yshift=-0.4em},
    ymin=0.95, ymax=2.1
]

\addplot[orange, mark=*, mark size=1.5pt, dash pattern=on 3pt off 5pt, line width=0.4pt] coordinates {
    (0.25,1.00) (0.5,1.12) (0.75,1.26) (1.0,1.40)
};

\addplot[cyan, mark=triangle*, mark size=1.5pt, dash pattern=on 3pt off 5pt, line width=0.4pt] coordinates {
    (0.25,1.08) (0.5,1.22) (0.75,1.38) (1.0,1.60)
};

\addplot[blue, mark=square*, mark size=1.3pt, dash pattern=on 3pt off 5pt, line width=0.4pt] coordinates {
    (0.25,1.15) (0.5,1.30) (0.75,1.49) (1.0,1.73)
};

\addplot[gray, mark=pentagon*, mark size=1.5pt, dash pattern=on 3pt off 5pt, line width=0.4pt] coordinates {
    (0.25,1.25) (0.5,1.50) (0.75,1.70) (1.0,2.00)
};

\addplot[pink, mark=diamond*, mark size=1.5pt, dash pattern=on 3pt off 5pt, line width=0.4pt] coordinates {
    (0.25,1.10) (0.5,1.28) (0.75,1.45) (1.0,1.68)
};


\nextgroupplot[
    ylabel=Norm. Exec.,
    xtick={0.25, 0.5, 0.75, 1.0},
    xmin=0.2, xmax=1.05,
    xticklabels={128, 256, 512, 1024},
    xticklabel style={
        rotate=25, anchor=east, font=\scriptsize,
        xshift=0.4em, yshift=-0.4em
    },
    ymin=0.95, ymax=2.2
]

\addplot[orange, mark=*, mark size=1.5pt, dash pattern=on 3pt off 5pt, line width=0.4pt] coordinates {
   (0.25,1.00) (0.5,1.17) (0.75,1.27) (1.0,1.55)
};

\addplot[cyan, mark=triangle*, mark size=1.5pt, dash pattern=on 3pt off 5pt, line width=0.4pt] coordinates {
     (0.25,1.06) (0.5,1.22) (0.75,1.35) (1.0,1.68)
   
};

\addplot[blue, mark=square*, mark size=1.3pt, dash pattern=on 3pt off 5pt, line width=0.4pt] coordinates {
    (0.25,1.12) (0.5,1.33) (0.75,1.58) (1.0,1.82)
};

\addplot[gray, mark=pentagon*, mark size=1.5pt, dash pattern=on 3pt off 5pt, line width=0.4pt] coordinates {
    (0.25,1.20) (0.5,1.48) (0.75,1.72) (1.0,2.10)
};

\addplot[pink, mark=diamond*, mark size=1.5pt, dash pattern=on 3pt off 5pt, line width=0.4pt] coordinates {
    (0.25,1.08) (0.5,1.28) (0.75,1.50) (1.0,1.72)
};

\nextgroupplot[
     ylabel=Norm. Exec.,
    xtick={0.25, 0.5, 0.75, 1.0},
    xmin=0.2, xmax=1.05,
    xticklabels={128, 256, 512, 1024},
    xticklabel style={rotate=25, anchor=east, font=\scriptsize, xshift=0.4em,
    yshift=-0.4em},
    ymin=0.95, ymax=2.1
]

\addplot[orange, mark=*, mark size=1.5pt, dash pattern=on 3pt off 5pt, line width=0.4pt] coordinates {
    (0.25,1.00) (0.5,1.12) (0.75,1.25) (1.0,1.45)
};

\addplot[cyan, mark=triangle*, mark size=1.5pt, dash pattern=on 3pt off 5pt, line width=0.4pt] coordinates {
    (0.25,1.03) (0.5,1.18) (0.75,1.35) (1.0,1.55)
};

\addplot[blue, mark=square*, mark size=1.3pt, dash pattern=on 3pt off 5pt, line width=0.4pt] coordinates {
    (0.25,1.10) (0.5,1.27) (0.75,1.45) (1.0,1.66)
};

\addplot[gray, mark=pentagon*, mark size=1.5pt, dash pattern=on 3pt off 5pt, line width=0.4pt] coordinates {
    (0.25,1.18) (0.5,1.42) (0.75,1.65) (1.0,2.00)
};

\addplot[pink, mark=diamond*, mark size=1.5pt, dash pattern=on 3pt off 5pt, line width=0.4pt] coordinates {
    (0.25,1.05) (0.5,1.22) (0.75,1.38) (1.0,1.58)
};

\nextgroupplot[
    ylabel=Norm. Exec.,
    xtick={0.25, 0.5, 0.75, 1.0},
    xmin=0.2, xmax=1.05,
    xticklabels={128, 256, 512, 1024},
    xticklabel style={rotate=25, anchor=east, font=\scriptsize, xshift=0.4em,
    yshift=-0.4em},
    ymin=0.95, ymax=2.5
]

\addplot[orange, mark=*, mark size=1.5pt, dash pattern=on 3pt off 5pt, line width=0.4pt] coordinates {
    (0.25,1.00) (0.5,1.20) (0.75,1.35) (1.0,1.72)
};

\addplot[cyan, mark=triangle*, mark size=1.5pt, dash pattern=on 3pt off 5pt, line width=0.4pt] coordinates {
    (0.25,1.10) (0.5,1.28) (0.75,1.5) (1.0,1.85)
   
};

\addplot[blue, mark=square*, mark size=1.3pt, dash pattern=on 3pt off 5pt, line width=0.4pt] coordinates {
    (0.25,1.18) (0.5,1.40) (0.75,1.68) (1.0,2.05)
};

\addplot[gray, mark=pentagon*, mark size=1.5pt, dash pattern=on 3pt off 5pt, line width=0.4pt] coordinates {
    (0.25,1.30) (0.5,1.58) (0.75,1.82) (1.0,2.35)
};

\addplot[pink, mark=diamond*, mark size=1.5pt, dash pattern=on 3pt off 5pt, line width=0.4pt] coordinates {
   (0.25,1.12) (0.5,1.35) (0.75,1.60) (1.0,1.95)
    
};

\end{groupplot}

\node at
  ($(group c2r1.north east)!0.5!(group c3r1.north west) + (1.7cm,0.3cm)$)
{%
  \scriptsize
  \begin{tabular}{@{\hspace{3pt}}ccccc@{\hspace{3pt}}}
    \tikz{
      \draw[orange, dash pattern=on 3pt off 5pt, line width=0.4pt]
        (0,0) -- (0.55,0);
      \draw[orange, fill=orange]
        (0.275,0) circle (1.2pt);
    }~\name
    &
    \tikz{
      \draw[cyan, dash pattern=on 3pt off 5pt, line width=0.4pt]
        (0,0) -- (0.55,0);
      \path[draw=cyan, fill=cyan]
        (0.275,0.10) -- (0.215,-0.05) -- (0.335,-0.05) -- cycle;
    }~\emph{DLFM}
    &
    \tikz{
      \draw[blue, dash pattern=on 3pt off 5pt, line width=0.4pt]
        (0,0) -- (0.55,0);
      \path[draw=blue, fill=blue]
        (0.225,-0.07) rectangle (0.325,0.07);
    }~\emph{3QTM}~\cite{shen2023thermal1}
    &
    \tikz{
      \draw[gray, dash pattern=on 3pt off 5pt, line width=0.4pt]
        (0,0) -- (0.55,0);
      \path[draw=gray, fill=gray]
        (0.275,0.11) --
        (0.215,0.02) --
        (0.235,-0.09) --
        (0.315,-0.09) --
        (0.335,0.02) -- cycle;
    }~\emph{DNaPE}~\cite{mohammed20233d}
    &
    \tikz{
      \draw[pink, dash pattern=on 3pt off 5pt, line width=0.4pt]
        (0,0) -- (0.55,0);
      \path[draw=pink, fill=pink]
        (0.275,0.10) --
        (0.215,0) --
        (0.275,-0.10) --
        (0.335,0) -- cycle;
    }~\emph{NeuroTAP}~\cite{pandey2024neurotap}
  \end{tabular}
};

\end{tikzpicture}
\vspace{-10pt}
    \caption{\small Normalized performance of \name~ across LFMs under varying input sequence lengths, with values referenced to \name~ at $L=128$.} 
    \label{fig:einput}
\end{figure*}
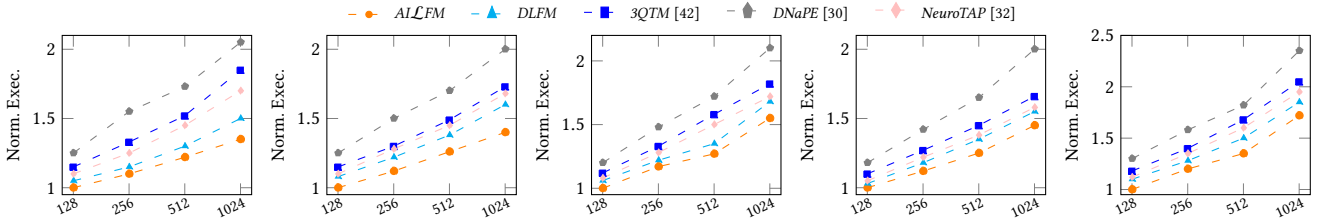

\subsection{Learning Policy}

We design a learning policy using a fully connected Neural Network (NN) with topology optimized using Neural Architecture Search (NAS)~\cite{zela2019understanding} to balance model complexity and performance. To enable uncertainty estimation, we integrate Monte Carlo~(MC) Dropout\cite{gal2016dropout}, which allows the agent to selectively query the Oracle based on its confidence in migration decisions. The optimal architecture identified by NAS consists of three hidden layers containing 64, 32, and 32 neurons,
each utilizing the ReLU activation function. 
The output layer comprises five neurons. Four identify the currently active LLM kernel through one-hot encoding and one responsible for regressing the predicted migration utility $\mathcal{I}$.

\subsubsection{Uncertainty Estimation with MC Dropout}

We employ MC Dropout~\cite{gal2016dropout} during inference to quantify the agent's confidence in its decisions. We estimate predictive uncertainty by incorporating dropout layers after fully connected layers in our NN model and keeping them active during inference. 

For an given input state $s$, we perform $\mathcal{N}$ stochastic forward passes with dropout enabled, yielding outputs $\{\hat{y}_1, \hat{y}_2, \dots, \hat{y}_N\}$. The mean prediction $\mu$ and predictive variance $\sigma^2$ are then computed, with $\sigma^2$ serving as the uncertainty estimate $u(s)$. Higher variance indicates greater uncertainty.

\subsubsection{Selective Querying Mechanism}

We define an uncertainty threshold $\tau$, determined empirically based on validation data, to decide when the agent should query the Oracle. The equation below formalizes the decision of an agent.

\begin{small}
\begin{equation}
\label{eq:decision_rule}
a =
\begin{cases}
\pi_\theta(s), & \text{if } u(s) \leq \tau, \\
\pi^*(s), & \text{if } u(s) > \tau,
\end{cases}
\end{equation}
\end{small}

where $\pi_\theta(s)$ is the agent's policy and $\pi^*(s)$ is the Oracle's policy. If the uncertainty $u(s)$ is below the threshold $\tau$, the agent proceeds with its own decision; otherwise, it queries the Oracle for the optimal action.

\subsubsection{Policy Training and Update}

We train the agent’s policy using a combination of autonomous and Oracle-guided actions. During training, we focus on reducing uncertainty in states where the agent previously queried the Oracle. The equation below defines the loss function $\mathcal{L}(\theta)$.

\begin{small}
\begin{equation}
\mathcal{L}(\theta) = \sum_{(s, a^*) \in \mathcal{D}_\text{Oracle}} \ell_{\text{sup}}(\pi_\theta(s), a^*) + \lambda \sum_{(s, a) \in \mathcal{D}_\text{Agent}} \ell_{\text{self}}(\pi_\theta(s), a),
\end{equation}
\end{small}

where $\mathcal{D}_\text{Oracle}$ represents states where the Oracle was queried, $\mathcal{D}_\text{Agent}$ contains autonomous decisions, $\ell_{\text{sup}}$ is the supervised loss (e.g., cross-entropy), $\ell_{\text{self}}$ is the self-imitation loss, and $\lambda$ balances the two. Emphasizing $\ell_{\text{sup}}$ helps the agent mimic the Oracle in uncertain states, while $\ell_{\text{self}}$ reinforces successful active actions to build confidence in similar scenarios.

\section{Experiment}
We validate our approach using interval thermal simulations on the state-of-the-art {\em CoMeT} simulator~\cite{siddhu2022comet}, which models a 3D S-NUCA system with 64 x86 cores and 128 memory banks. The 3D S-NUCA system employs a 4x4x4 core topology with XYZ routing and a 4x4x8 memory bank topology. The core topology incorporates a 3D NoC with XYZ routing, while the memory bank topology uses a synchronized bus. The cores and memory banks have dimensions of 3.414\,mm by 3.414\,mm. Each core operates at a maximum clock frequency of 3\,GHz and features a private 32\,KB L1 data cache, a private 32\,KB L1 instruction cache, and a shared 1\,MB S-NUCA cache per core, resulting in a total 64\,MB on-chip shared LLC. We use a 64\,GB HBM memory with a bandwidth of 25.6\,GB/s.  


The workloads include standalone implementations of five LFM models: {\em BERT-base}~\cite{kenton2019bert} on SQuAD v1.1, {\em ViT-base}~\cite{dosovitskiy2020image} on ImageNet-1K, {\em LLaMA3.2-1B}~\cite{touvron2023llama} on WikiText-103, {\em Gemma-2B}~\cite{gemma2024improving} on C4 dataset, and {\em DeepSeek-2.4B}~\cite{guo2025deepseek} on OpenWebText2.
We compare \name~against state-of-the-art thermal management and scheduling techniques for 3D-stacked architectures, including {\em 3QTM}~\cite{shen2023thermal1}, a reinforcement learning-based core and memory coordination method for thermal management; {\em NeuroTAP}~\cite{pandey2024neurotap}, an analytical model-based thermal management approach for 3D DRAM; {\em 3D-DNaPE}~\cite{mohammed20233d}, a scheduling technique that migrates tasks to the coldest adjacent inactive cores; and {\em DLFM}, a direct learning-based baseline without Oracle guidance.

\makeatletter
\renewcommand\thesubsubsection{\arabic{section}.\arabic{subsubsection}}
\@addtoreset{subsubsection}{section}
\makeatother


\subsubsection{\textbf{Impact of LFM Input Size on System Performance.}}
We explpore the impact of varying input sequence lengths of $L = 128, 256, 512, 1024$ tokens on the execution performance of diverse LFM workloads under a thermal threshold of $T_{th} = 80$\,\textcelsius. The evaluated models include {\em ViT-base}, {\em BERT-base}, {\em LLaMA3.2-1B}, {\em Gemma-2B}, and {\em DeepSeek-2.4B}. Figure~\ref{fig:einput} reports execution performance normalized to \name{} at $L{=}128$. Across all models and input sizes, \name~ outperforms baselines, with gains becoming more pronounced as input length increases, highlighting its robustness under elevated computational and thermal stress. Specifically, as sequence length scales, \name~maintains robust performance relative to baselines. E.g., on {\em DeepSeek-2.4B}, \name~achieves a notable performance advantage, outperforming the closest competitor, {\em DNaPE}, by approximately 35\% at 1024 tokens. 



\begin{figure} [t]
\begin{tikzpicture}
\footnotesize
\begin{groupplot}[
    group style={
        group size=2 by 2, 
        vertical sep=1.5cm, 
    },
    width=5cm,
    height=3.5cm,
    ylabel style={align=center},
    xlabel style={yshift=-5pt},
    every axis plot/.append style={line width=1pt},
]


\nextgroupplot[
    ylabel=Norm. Exec.,
    xtick={0.2, 0.4, 0.6, 0.8, 1}, 
    xmin=0.1, xmax=1.1,
    xticklabels={ViT-base, BERT-base, LLaMA3.2 1B,  Gemma-2B,DeepSeek-2.4B}, 
    xticklabel style={rotate=25, anchor=east, font=\scriptsize,xshift=0.3em}, 
    bar width=1.6pt,
    ybar,
    ymin=0.95, ymax=1.35 
]
\addplot[orange, fill=orange!30, postaction={pattern=north east lines}] coordinates {(0.2,1) (0.4,1) (0.6,1) (0.8,1) (1,1)};
\addplot[cyan, fill=cyan!30, postaction={pattern=north west lines}] coordinates {(0.2,1.01) (0.4,1.03) (0.6,1.04) (0.8,1.06) (1,1.09)};
\addplot[blue, fill=blue!30, postaction={pattern=horizontal lines}] coordinates {(0.2,1.08) (0.4,1.09) (0.6,1.12) (0.8,1.11) (1,1.10)};
\addplot[gray, fill=gray!30, postaction={pattern=crosshatch}] coordinates {(0.2,1.14) (0.4,1.13) (0.6,1.25) (0.8,1.30) (1,1.31)};
\addplot[pink, fill=pink!30] coordinates {(0.2,1.07) (0.4,1.09) (0.6,1.11) (0.8,1.13) (1,1.13)};
\draw[densely dotted, thick, line width=1pt, |->] 
             (axis cs:1.03, 1.31) -- (axis cs:0.92, 1.0);
\node[red] at (0.95,1.19) {$\uparrow$31\%};


\nextgroupplot[
    ylabel=Norm. Exec.,
    xtick={0.2, 0.4, 0.6, 0.8, 1}, 
    xmin=0.1, xmax=1.1,
    xticklabels={ViT-base, BERT-base, LLaMA3.2 1B,  Gemma-2B,DeepSeek-2.4B}, 
    xticklabel style={rotate=25, anchor=east, font=\scriptsize,xshift=0.3em}, 
    bar width=1.6pt,
    ybar,
    ymin=0.95, ymax=1.3 
]
\addplot[orange, fill=orange!30, postaction={pattern=north east lines}] coordinates {(0.2,1) (0.4,1) (0.6,1) (0.8,1) (1,1)};
\addplot[cyan, fill=cyan!30, postaction={pattern=north west lines}] coordinates {(0.2,1.05) (0.4,1.08) (0.6,1.08) (0.8,1.09) (1,1.10)};
\addplot[blue, fill=blue!30, postaction={pattern=horizontal lines}] coordinates {(0.2,1.06) (0.4,1.13) (0.6,1.13) (0.8,1.13) (1,1.15)};
\addplot[gray, fill=gray!30, postaction={pattern=crosshatch}] coordinates {(0.2,1.16) (0.4,1.17) (0.6,1.25) (0.8,1.27) (1,1.30)};
\addplot[pink, fill=pink!30] coordinates {(0.2,1.09) (0.4,1.12) (0.6,1.13) (0.8,1.09) (1,1.12)};
\draw[densely dotted, thick, line width=1pt, |->] 
            (axis cs:1.03, 1.30) -- (axis cs:0.92, 1.0);
\node[red] at (0.95,1.19) {$\uparrow$30\%};

\end{groupplot}

\node at ($(group c1r1.north east)!0.5!(group c2r1.north west) + (0cm, 0.3cm)$) {
    \begin{tabular}{@{\hspace{-10pt}}ccccc@{\hspace{-10pt}}}
        \tikz\draw[draw=orange, fill=orange!30, pattern=north east lines, pattern color=orange] (0,0) rectangle (5pt,5pt); \scriptsize \name &
        \tikz\draw[draw=cyan, fill=cyan!30, pattern=north west lines, pattern color=cyan] (0,0) rectangle (5pt,5pt); \scriptsize {\em DLFM} &
        \tikz\draw[draw=blue, fill=blue!30, pattern=horizontal lines, pattern color=blue] (0,0) rectangle (5pt,5pt); \scriptsize {\em 3QTM}~\cite{shen2023thermal1} &
        \tikz\draw[draw=gray, fill=gray!30, pattern=crosshatch, pattern color=gray] (0,0) rectangle (5pt,5pt); \scriptsize {\em DNaPE}~\cite{mohammed20233d} &
        \tikz\draw[draw=pink, fill=pink!30] (0,0) rectangle (5pt,5pt); \scriptsize {\em NeuroTAP}~\cite{pandey2024neurotap} \\
    \end{tabular}
};

\node at ($(group c1r1.south) + (0,-1cm)$) {\scriptsize (a) $T_{th}=75$~\textcelsius};
\node at ($(group c2r1.south) + (0,-1cm)$) {\scriptsize (b) $T_{th}=85$~\textcelsius};

\end{tikzpicture}
\vspace{-23pt}
    \caption{\small Normalized performance of \name~ versus baselines.}

    \label{fig:perf}
\end{figure}
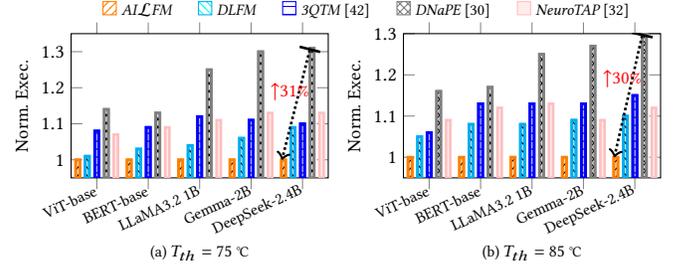

\vspace{-5pt}
\subsubsection{\textbf{Performance Comparison Across LFM Workloads and Thermal Thresholds.}}
We compare \name's performance against the baselines across various LFMs, under thermal thresholds ($T_{th}$) of 75\,\textcelsius and 85\,\textcelsius~with an input size of $L=256$. Figure~\ref{fig:perf} illustrates that \name~consistently achieves superior performance relative to baselines across all LFMs and thermal thresholds. Specifically, under $T_{th}=75$\,\textcelsius, \name~significantly outperforms other approaches, achieving up to 31\% better performance versus {\em DNaPE}. 
This trend consistently holds across all LFMs.


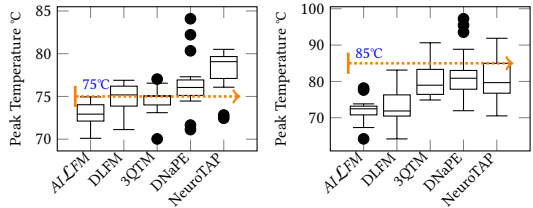
\begin{figure} [t]
    \centering
    \subfigure[Temperature threshold 75\textcelsius]{
    \begin{tikzpicture}
		\scriptsize
		\begin{axis}[
        width = 0.5\columnwidth,
        height = \chartheight, 
        boxplot/draw direction=y,
        ylabel={Peak Temperature~\textcelsius},
        xtick={1,2,3,4,5},
        xticklabels={{\name}, {DLFM}, {3QTM}, {DNaPE}, {NeuroTAP}}, 
        xticklabel style={rotate=45, anchor=east, font=\scriptsize} 
    ]

		\addplot+[mark options={draw=black}, mark=*, color=black,  solid, boxplot] table[ x = Index, y = AILFM, ]{dat/11aaaaa.txt};
		\addplot+[mark options={draw=black}, mark=*, color=black,  solid, boxplot] table[ x = Index, y = DLFM, ]{dat/11aaaaa.txt};
		\addplot+[mark options={draw=black}, mark=*, color=black,  solid, boxplot] table[ x = Index, y = 3QTM, ]{dat/11aaaaa.txt};
		\addplot+[mark options={draw=black}, mark=*, color=black,  solid, boxplot] table[ x = Index, y = DNaPE, ]{dat/11aaaaa.txt};
  \addplot+[mark options={draw=black}, mark=*, color=black,  solid, boxplot] table[ x = Index, y = NeuroTAP, ]{dat/11aaaaa.txt};
  \addplot [densely dotted, orange, line width=1pt,, |->] coordinates {(0.5,75) (5.5,75)};
  
  \node[below,blue] at (1.17,77.7) {75\textcelsius};

		\end{axis}
		\end{tikzpicture}}  
    \subfigure[Temperature threshold 85\textcelsius]{
    \begin{tikzpicture}
		\scriptsize
		\begin{axis}[
		width = 0.5\columnwidth,
		height=\chartheight,
		boxplot/draw direction=y,
		ylabel={Peak Temperature~\textcelsius},
		xtick={1,2,3,4,5},
            xticklabels={{\name}, {DLFM}, {3QTM}, {DNaPE}, {NeuroTAP}}, 
            xticklabel style={rotate=45, anchor=east, font=\scriptsize} 
        ],

		\addplot+[mark options={draw=black}, mark=*, color=black,  solid, boxplot] table[ x = Index, y = AILFM, ]{dat/aaaaa.txt};
		\addplot+[mark options={draw=black}, mark=*, color=black,  solid, boxplot] table[ x = Index, y = DLFM, ]{dat/aaaaa.txt};
		\addplot+[mark options={draw=black}, mark=*, color=black,  solid, boxplot] table[ x = Index, y = 3QTM, ]{dat/aaaaa.txt};
		\addplot+[mark options={draw=black}, mark=*, color=black,  solid, boxplot] table[ x = Index, y = DNaPE, ]{dat/aaaaa.txt};
  \addplot+[mark options={draw=black}, mark=*, color=black,  solid, boxplot] table[ x = Index, y = NeuroTAP, ]{dat/aaaaa.txt};
  \addplot [densely dotted, orange, line width=1pt,, |->] coordinates {(0.5,85) (5.5,85)};

   \node[below,blue] at (1.17,90.7) {85\textcelsius};

		\end{axis}
		\end{tikzpicture}
  }
\vspace{-7pt}  
    \caption{\small $T_{peak}$ comparison between baselines of 3D S-NUCA.}
    
    \label{fig:temp}

\end{figure}

\subsubsection{\textbf{Peak Temperature Analysis Across Thermal Thresholds.}}
Figure~\ref{fig:temp} shows the comparison of peak temperatures across baselines, highlighting the superior thermal management of \name~versus other baselines. On average, \name~maintains a lower peak temperature, with an overall mean of approximately 72.5\,°C in the 75\,°C threshold and 73.2\,°C in the 85\,°C  threshold, consistently outperforming other methods. 
{\em NeuroTAP} demonstrates suboptimal thermal performance, with peak temperatures often ranging between 77\,°C and 80\,°C when $T_{th}=$\,75\textcelsius, due to its analytical basis and limited adaptability to LFM-induced workload variability.

\subsubsection{\textbf{Inference Overhead}}
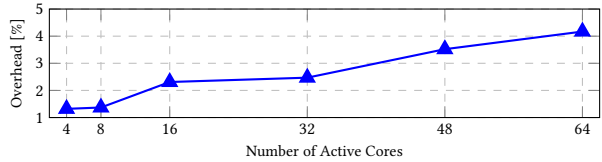
\begin{figure}[t]
\scriptsize
\centering
\begin{tikzpicture}
    \begin{axis}[
        width=0.5\textwidth,
        height=0.17\textwidth,
        xlabel={Number of Active Cores},
        ylabel={Overhead [$\%$]},
        xmin=2, xmax=66,
        ymin=1, ymax=5,
        xtick={4, 8, 16, 32, 48, 64},
        ytick={1, 2, 3, 4, 5}, 
        grid=both,
        minor grid style={dotted},
        major grid style={dashed},
        mark options={solid},
    ]
    \addplot[
        mark=triangle*,
        mark size=3pt,
        thick,
        blue, 
    ] coordinates {
        (4, 1.32)
        (8, 1.37)
        (16, 2.31)
        (32, 2.47)
        (48, 3.52)
        (64, 4.17)
    };
    \end{axis}
\end{tikzpicture}
\vspace{-10pt}
\caption{\small Run-time overhead with increasing active cores.}
\label{fig:overhead_vs_cores}
\end{figure}

Figure~\ref{fig:overhead_vs_cores} shows the sensitivity and scalability analysis of \name's scheduling epoch size and associated run-time overhead. Leveraging MC Dropout and a lightweight neural network, \name~maintains a run-time overhead of less than 5\%, even under a fully loaded system.

\begin{table}[t]
\centering
\caption{\small Scalability across cache organizations and 3D topologies.}
\vspace{-10pt}
\resizebox{0.50\textwidth}{!}{
\begin{tabular}{lccc}
\toprule
Architecture 
& Norm. Time $\downarrow$ 
& $T_{\text{peak}}$ [$^\circ$C] $\downarrow$ 
& speedup $\uparrow$ \\
\midrule
2D planar S-NUCA (8$\times$8, 64 cores) 
  & 1.00 
  & \fs73.5 
  & - \\
3D cores w/o S-NUCA (4$\times$4$\times$4, 64 cores)
  & 0.88
  & 77.1
  & 1.14$\times$ \\
3D S-NUCA (4$\times$4$\times$4, 64 cores) 
  & \nd0.67 
  & \nd75.2 
  & \nd1.49$\times$ \\
3D cores w/o S-NUCA (6$\times$6$\times$6, 216 cores)
  & 0.79
  & 81.6
  & 1.27$\times$ \\
3D S-NUCA (6$\times$6$\times$6, 216 cores) 
  & \fs0.61 
  & 83.4 
  & \fs1.64$\times$ \\
\bottomrule
\end{tabular}
}
\label{tab:scalability-3d}
\end{table}

\subsubsection{\textbf{Scalability Across Cache Organizations and 3D Topologies}}
Table~\ref{tab:scalability-3d} shows that \name{} scales reliably across architectures when running {\em DeepSeek-2.4B}. Relative to the 2D planar baseline, the 4$\times$4$\times$4 3D design without S-NUCA already improves performance (1.14$\times$ speedup) due to increased parallelism, despite higher thermal density. Adding S-NUCA further amplifies these gains by restoring locality, reducing normalized time to 0.67 and achieving a 1.49$\times$ speedup with similar peak temperatures. Scaling to a 216-core 6$\times$6$\times$6 stack increases thermal stress, as expected in deeper 3D systems, yet yields the highest throughput; the S-NUCA variant reaches a 1.64$\times$ speedup. 

\vspace{-3pt}

\subsubsection{\textbf{Oracle-Guided Learning vs. Direct Policy Learning}}

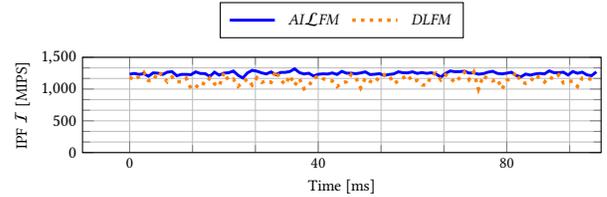
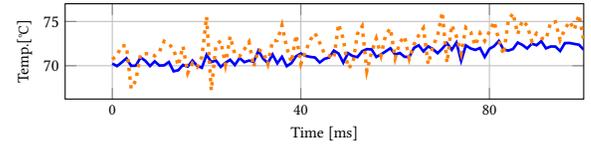
\begin{figure} [!t]
\scriptsize
\subfigure[Performance]{
\begin{tikzpicture}
    \pgfplotsset{compat=1.17}
    \begin{axis}[
            width  = \columnwidth,
            height=\ctheight*0.77, 
            ylabel= {\scriptsize IPF $\mathcal{I}$ [MIPS]},
            xlabel = {Time [ms]},
            xmax = 100,
            ymax = 1500,
            ymin= 0,
            grid=both,
            minor tick num=2,
            xtick= {0,40,80,120,160,200},
            xticklabels= {\scriptsize 0,40,80,120,160,200},
             legend columns=4,
             legend cell align=left,
             legend style={
                at={(0.5,1.2)},
                 anchor=south,
                 column sep=1ex},
            ]

            \addlegendentry{\scriptsize \name};
            \addlegendentry{\scriptsize \em DLFM};
               
            \addplot [blue,line width=1.1pt] table[x expr=\coordindex,y=Value, col sep=comma] {data/maxFreqcore5.txt};
            \addplot [orange, dotted,line width=1.3pt] table[x expr=\coordindex,y=Value, col sep=comma] {data/maxFreqcore10.txt};
        \end{axis} 
    \end{tikzpicture}
} 
\subfigure[Temperature]{
\hspace{0.23cm}
    \begin{tikzpicture}
    \pgfplotsset{compat=1.17}
    \begin{axis}[
            width  = \columnwidth,
            height=\ctheight*0.77, 
            ylabel= {\scriptsize Temp.[\textcelsius]},
            xlabel = {Time [ms]},
            grid=both,
            xmax = 100,
            ymax = 77,
             xtick= {0,40,80,120,160,200},
            xticklabels= {0,40,80,120,160,200},
            ]
    
            \addplot [blue,line width=1.01pt] table[x expr=\coordindex,y=Value] {data/tspcore5.txt};
            \addplot [orange, dotted,line width=1.3pt] table[x expr=\coordindex,y=Value] {data/tspcore10.txt};

        \end{axis} 
    \end{tikzpicture}
}
\vspace{-0.5cm}
\caption {\small Comparison of~\name~ and DLFM shows \name~delivers higher IPS and lower, more stable temperatures via its MoGPR.}
\label{fig:motiresults}
\vspace{-0.5cm}
\end{figure}

 We present an illustrative example comparing the migration decisions of \name~and {\em DLFM} during the execution of the attention kernel from the LFM {\em BERT-base} model when $T_{th} = $ 75\,\textcelsius. Figure~\ref{fig:motiresults} shows that \name~consistently achieves a more stable and higher $\mathcal{I}$ while maintaining lower and more stable temperatures. This improvement stems from leveraging expert-labeled data from MoGPR-based Oracle demonstrations to train a kernel-aware scheduling policy. By explicitly modeling the performance–thermal tradeoffs for each kernel type, \name~learns a policy that avoids the pitfalls of direct learning, such as catastrophic forgetting, which commonly arises when a unified model trains over heterogeneous kernels.

 \begin{table}[t]
\centering
\caption{\small Oracle upper bound versus \name{}.
}
\vspace{-7pt}
\resizebox{0.47\textwidth}{!}{
\begin{tabular}{lcccc}
\toprule
Method 
& Accuracy [\%] $\uparrow$
& Norm. Time $\downarrow$ 
& $T_{\text{peak}}$ [$^\circ$C] $\downarrow$
& Overhead [\%] $\downarrow$ \\
\midrule
\textbf{Oracle (MoGPR, upper bound)} 
  & \fs97.3 
  & 1
  & \fs74.3
  & 24.6\\
\textbf{\name{} (ours)} 
  & \nd96.7 
  & \fs0.13
  & 75.1 
  & \fs3.2 \\
\bottomrule
\end{tabular}
}
\label{tab:oracle-upper-bound}
\end{table}

\subsubsection{\textbf{Oracle Upper Bound Comparison}}
Table~\ref{tab:oracle-upper-bound} shows that the MoGPR Oracle offers only slight gains in accuracy and $T_{\text{peak}}$ over \name, but at a substantial cost: each epoch requires full MoGPR inference, resulting in 24.6\% overhead. In contrast, \name{} achieves near-oracle accuracy with only 3.2\% overhead and significantly faster execution, making it far more practical for real-time use.

\begin{table}[t]
\centering
\caption{\small Utility prediction accuracy with different models.}
\vspace{-10pt}
\resizebox{0.43\textwidth}{!}{
\begin{tabular}{lccc}
\toprule
Model & Utility MAE$\downarrow$ & Exec. Time (s)$\downarrow$ & Queries/Epoch $\downarrow$\\
\midrule
Random Forest & 0.42 & \nd103.7 & 0.21 \\
MLP & \nd0.39 & 150.9 & \nd0.18 \\
MoGPR (ours) & \fs0.31 & \fs79.1 & \fs0.11 \\
\bottomrule
\end{tabular}
}
\label{tab:predictive-models}
\end{table}

\subsubsection{\textbf{Utility Prediction Accuracy}}
We compare predictive models (RF, MLP, MoGPR) by their ability to estimate migration utility, defined as performance gain minus cold-start overhead. 
We report \emph{Utility MAE}, the mean absolute error between predicted and measured utility. 
A lower MAE enables more accurate migration decisions. 
Table~\ref{tab:predictive-models} shows that  MoGPR achieves the lowest Utility MAE, fastest execution time, and overall best scheduling.

\subsubsection{\textbf{Uncertainty Threshold $\tau$ Sensitivity}}
We vary the $\tau$ in the \name ~module to study its effect on Oracle queries and thermal safety. 
Table~\ref{tab:tau-sensitivity} shows that lower $\tau$ triggers more Oracle queries, while higher $\tau$ reduces queries but risks occasional thermal violations. 
Safe overrides are fallback actions (DVFS or migration) triggered when $T_{\text{peak}}$ nears $T_{\text{th}}$ to avoid violations.
The mid-range setting ($\tau=0.15$) achieves the best trade-off, with only 0.12 queries/epoch, 0.07\% violations, and minimal runtime overhead. 

\begin{table}[t]
\centering
\caption{\small Sensitivity of the uncertainty threshold $\tau$. 
}
\vspace{-10pt}
\resizebox{0.41\textwidth}{!}{
\begin{tabular}{lccc}
\toprule
$\tau$ & Safe overrides (\%)$\downarrow$ & Violations (\%)$\downarrow$ & Overhead (\%)$\downarrow$ \\
\midrule
0.05 & 7.9  & \fs0.07 & 2.5 \\
0.10 & 6.1  & 0.09 & 1.0 \\
0.15 & \fs{5.4} & \fs{0.07} & \fs{0.2} \\
0.20 & \nd5.1  & 0.23 & \nd1.6 \\
0.30 & 4.8  & 0.92 & 3.2 \\
\bottomrule
\end{tabular}
}
\label{tab:tau-sensitivity}
\end{table}



\begin{table}[t]
\centering
\caption{\small Ablation of kernel-aware modeling and active querying.}
\vspace{-10pt}
\resizebox{0.43\textwidth}{!}{
\begin{tabular}{lccc}
\toprule
Method 
& Acc. [\%] $\uparrow$ 
& Norm. Time $\downarrow$
& $T_{\text{peak}}$ [$^\circ$C] $\downarrow$ \\
\midrule
\textbf{\name{} (ours)} 
  & \fs96.7 
  & \fs0.77 
  & \fs75.1 \\
\fork~  Kernel-aware 
  & 90.8 
  & \nd0.92 
  & \nd79.9 \\
\fork~ Active querying 
  & \nd91.1 
  & 0.95 
  & 81.4 \\
DLFM (direct learning) 
  & 90.3 
  & 1.00 
  & 82.7 \\
\bottomrule
\end{tabular}
}
\label{tab:core-ablation}
\end{table}


\subsubsection{\textbf{Kernel-Aware and Active Query Ablation}}
Table~\ref{tab:core-ablation} shows that removing kernel-aware modeling or active querying significantly degrades all metrics. Without kernel-awareness, accuracy drops sharply and $T_{\text{peak}}$ rises to 79.9$^\circ$C, while disabling active querying further slows execution and increases thermal stress. These results confirm that both kernel-specific scheduling and uncertainty-driven querying are key to \name{}’s efficiency and thermal stability.

\section{Conclusion}
This paper introduces \name, a new AIL-based framework for thermal- and kernel-aware LFM inference on 3D-stacked S-NUCA systems. 
By leveraging the heterogeneity of 3D S-NUCA systems and accounting for the kernel-specific characteristics of LFMs, \name~first performs a one-time offline MoGPR-based Oracle demonstration to generate reliable migration utility knowledge. 
It then distills this knowledge into a lightweight model through AIL. 
\name{} yields up to 31\% higher execution performance with under 5\% run-time overhead, even under fully loaded conditions, and maintains these gains as system size increases, scaling reliably from 2D planar chips to large 3D topologies.


\begin{acks}
This work was funded under the EU Horizon Europe COIN-3D project (Grant No. 101159667).
\end{acks}


\bibliographystyle{acm}
\bibliography{ref}

@inproceedings{kuper2024quantitative,
  title={A Quantitative Analysis and Guidelines of Data Streaming Accelerator in Modern Intel Xeon Scalable Processors},
  author={Kuper, Reese and Jeong, Ipoom and Yuan, Yifan and others},
  booktitle={Proceedings of the 29th ACM International Conference on Architectural Support for Programming Languages and Operating Systems, Volume 2},
  pages={37--54},
  year={2024}
}

@inproceedings{liu2019optimizing,
  title={Optimizing $\{$CNN$\}$ model inference on $\{$CPUs$\}$},
  author={Liu, Yizhi and Wang, Yao and Yu, Ruofei and others},
  booktitle={2019 USENIX Annual Technical Conference (USENIX ATC 19)},
  pages={1025--1040},
  year={2019}
}

@inproceedings{perryman2023evaluation,
  title={Evaluation of xilinx versal architecture for next-gen edge computing in space},
  author={Perryman, Noah and Wilson, Christopher and George, Alan},
  booktitle={2023 IEEE aerospace conference},
  pages={1--11},
  year={2023},
  organization={IEEE}
}

@inproceedings{hazelwood2018applied,
  title={Applied machine learning at facebook: A datacenter infrastructure perspective},
  author={Hazelwood, Kim and Bird, Sarah and and others},
  booktitle={2018 IEEE international symposium on high performance computer architecture (HPCA)},
  pages={620--629},
  year={2018},
  organization={IEEE}
}

@inproceedings{wu2019machine,
  title={Machine learning at facebook: Understanding inference at the edge},
  author={Wu, Carole-Jean and Brooks, David and Chen, Kevin and Chen, Douglas and Choudhury, Sy and Dukhan, Marat and Hazelwood, Kim and Isaac, Eldad and Jia, Yangqing and Jia, Bill and others},
  booktitle={2019 IEEE international symposium on high performance computer architecture (HPCA)},
  pages={331--344},
  year={2019},
  organization={IEEE}
}

@article{dosovitskiy2020image,
  title={An image is worth 16x16 words: Transformers for image recognition at scale},
  author={Dosovitskiy, Alexey},
  journal={arXiv preprint arXiv:2010.11929},
  year={2020}
}

@inproceedings{he2016deep,
  title={Deep residual learning for image recognition},
  author={He, Kaiming and Zhang, Xiangyu and Ren, Shaoqing and Sun, Jian},
  booktitle={Proceedings of the IEEE conference on computer vision and pattern recognition},
  pages={770--778},
  year={2016}
}

@inproceedings{huh2005nuca,
  title={A NUCA substrate for flexible CMP cache sharing},
  author={Huh, Jaehyuk and Kim, Changkyu and Shafi, Hazim and Zhang, Lixin and Burger, Doug and Keckler, Stephen W},
  booktitle={ACM International Conference on Supercomputing 25th Anniversary Volume},
  pages={380--389},
  year={2005}
}

@article{merino2008sp,
  title={Sp-nuca: a cost effective dynamic non-uniform cache architecture},
  author={Merino, Javier and Puente, Valent{\'\i}n and Prieto, Pablo and Gregorio, Jos{\'e} {\'A}ngel},
  journal={ACM SIGARCH Computer Architecture News},
  volume={36},
  number={2},
  pages={64--71},
  year={2008},
  publisher={ACM New York, NY, USA}
}

@article{siddhu2022comet,
  title={CoMeT: An integrated interval thermal simulation toolchain for 2D, 2.5 D, and 3D processor-memory systems},
  author={Siddhu, Lokesh and Kedia, Rajesh and others},
  journal={ACM Transactions on Architecture and Code Optimization (TACO)},
  volume={19},
  number={3},
  pages={1--25},
  year={2022},
  publisher={ACM New York, NY},
  doi={https://doi.org/10.1145/3532185}
}

@article{hsieh2013thermal,
  title={Thermal-aware memory mapping in 3D designs},
  author={Hsieh, Ang-Chih and Hwang, TingTing},
  journal={ACM Transactions on Embedded Computing Systems (TECS)},
  volume={13},
  number={1},
  pages={1--22},
  year={2013},
  publisher={ACM New York, NY, USA},
  doi={https://doi.org/10.1145/2512457}
}

@inproceedings{lo2016thermal,
  title={Thermal-aware dynamic page allocation policy by future access patterns for Hybrid Memory Cube (HMC)},
  author={Lo, Wei-Hen and Liang, Kai-zen and Hwang, TingTing},
  booktitle={2016 Design, Automation \& Test in Europe Conference \& Exhibition (DATE)},
  pages={1084--1089},
  year={2016},
  organization={IEEE},
  url={https://ieeexplore.ieee.org/document/7459470}
}

@article{liu2022tb,
  title={TB-NUCA: A Temperature-Balanced 3D NUCA Based on Bayesian Optimization},
  author={Liu, Hanyan and Zhao, Yunping and Chen, Xiaowen and Li, Chen and Lu, Jianzhuang},
  journal={Electronics},
  volume={11},
  number={18},
  pages={2910},
  year={2022},
  publisher={MDPI}
}

@inproceedings{tsai2012thermal,
  title={Thermal-aware real-time task scheduling for three-dimensional multicore chip},
  author={Tsai, Ting-Hao and Chen, Ya-Shu},
  booktitle={Proceedings of the 27th Annual ACM Symposium on Applied Computing},
  year={2012}
}

@inproceedings{chaturvedi2014thermal,
  title={Thermal-aware task scheduling for peak temperature minimization under periodic constraint for 3D-MPSoCs},
  author={Chaturvedi, Vivek and Singh, Amit Kumar and Zhang, Wei and Srikanthan, Thambipillai},
  booktitle={2014 25nd IEEE International Symposium on Rapid System Prototyping},
  year={2014},
  organization={IEEE}
}

@inproceedings{meng2012optimizing,
  title={Optimizing energy efficiency of 3-D multicore systems with stacked DRAM under power and thermal constraints},
  author={Meng, Jie and Kawakami, Katsutoshi and Coskun, Ayse K},
  booktitle={Proceedings of the 49th Annual Design Automation Conference},
  pages={648--655},
  year={2012}
}

@article{kumar2017fighting,
  title={Fighting dark silicon: Toward realizing efficient thermal-aware 3-D stacked multiprocessors},
  author={Kumar, Sumeet S and Zjajo, Amir and van Leuken, Rene},
  journal={IEEE Transactions on Very Large Scale Integration (VLSI) Systems},
  volume={25},
  number={4},
  pages={1549--1562},
  year={2017},
  publisher={IEEE}
}

@article{siddhu2020leakage,
  title={Leakage-aware dynamic thermal management of 3D memories},
  author={Siddhu, Lokesh and Kedia, Rajesh and Panda, Preeti Ranjan},
  journal={ACM Transactions on Design Automation of Electronic Systems (TODAES)},
  year={2020},
  publisher={ACM New York, NY, USA}
}

@inproceedings{henkel2018dynamic,
  title={Dynamic resource management for heterogeneous many-cores},
  author={Henkel, J{\"o}rg and Teich, J{\"u}rgen and Wildermann, Stefan and Amrouch, Hussam},
  booktitle={2018 IEEE/ACM International Conference on Computer-Aided Design (ICCAD)},
  pages={1--6},
  year={2018},
  organization={IEEE}
}

@inproceedings{shen2023thermal,
  title={Thermal management for s-nuca many-cores via synchronous thread rotations},
  author={Shen, Yixian and Niknam, Sobhan and Pathania, Anuj and Pimentel, Andy D},
  booktitle={2023 Design, Automation \& Test in Europe Conference \& Exhibition (DATE)},
  pages={1--6},
  year={2023},
  organization={IEEE}
}

@article{zhang2017machine,
  title={Machine learning-based temperature prediction for runtime thermal management across system components},
  author={Zhang, Kaicheng and Guliani, Akhil and Ogrenci-Memik, Seda and Memik, Gokhan and Yoshii, Kazutomo and Sankaran, Rajesh and Beckman, Pete},
  journal={IEEE Transactions on parallel and distributed systems},
  volume={29},
  number={2},
  pages={405--419},
  year={2017},
  publisher={IEEE}
}

@article{pandey2024neurotap,
  title={NeuroTAP: Thermal and Memory Access Pattern-Aware Data Mapping on 3D DRAM for Maximizing DNN Performance},
  author={Pandey, Shailja and Panda, Preeti Ranjan},
  journal={ACM Transactions on Embedded Computing Systems},
  volume={23},
  number={6},
  pages={1--30},
  year={2024},
  publisher={ACM New York, NY}
}

@article{shen2023thermal1,
  title={Thermal management for 3d-stacked systems via unified core-memory power regulation},
  author={Shen, Yixian and Schreuders, Leo and Pathania, Anuj and Pimentel, Andy D},
  journal={ACM Transactions on Embedded Computing Systems},
  volume={22},
  number={5s},
  pages={1--26},
  year={2023},
  publisher={ACM New York, NY}
}

@inproceedings{sikal2023machine,
  title={Machine Learning-based Thermally-Safe Cache Contention Mitigation in Clustered Manycores},
  author={Sikal, Mohammed Bakr and Khdr, Heba and Rapp, Martin and Henkel, J{\"o}rg},
  booktitle={2023 60th ACM/IEEE Design Automation Conference (DAC)},
  pages={1--6},
  year={2023},
  organization={IEEE}
}

@article{guo2025deepseek,
  title={Deepseek-r1: Incentivizing reasoning capability in llms via reinforcement learning},
  author={Guo, Daya and Yang, Dejian and Zhang, Haowei and Song, Junxiao and Zhang, Ruoyu and Xu, Runxin and Zhu, Qihao and Ma, Shirong and Wang, Peiyi and Bi, Xiao and others},
  journal={arXiv preprint arXiv:2501.12948},
  year={2025}
}

@inproceedings{kenton2019bert,
  title={Bert: Pre-training of deep bidirectional transformers for language understanding},
  author={Kenton, Jacob Devlin Ming-Wei Chang and Toutanova, Lee Kristina},
  booktitle={Proceedings of naacL-HLT},
  volume={1},
  pages={2},
  year={2019},
  organization={Minneapolis, Minnesota}
}

@article{shen2023efficient,
  title={Efficient llm inference on cpus},
  author={Shen, Haihao and Chang, Hanwen and Dong, Bo and Luo, Yu and Meng, Hengyu},
  journal={arXiv preprint arXiv:2311.00502},
  year={2023}
}

@article{he2024inference,
  title={Inference performance optimization for large language models on cpus},
  author={He, Pujiang and Zhou, Shan and Huang, Wenhuan and Li, Changqing and Wang, Duyi and Guo, Bin and Meng, Chen and Gui, Sheng and Yu, Weifei and Xie, Yi},
  journal={arXiv preprint arXiv:2407.07304},
  year={2024}
}

@article{huang2006hotspot,
  title={HotSpot: A compact thermal modeling methodology for early-stage VLSI design},
  author={Huang, Wei and Ghosh, Shougata and Velusamy, Sivakumar and others},
  journal={IEEE Transactions on very large scale integration (VLSI) systems},
  volume={14},
  number={5},
  pages={501--513},
  year={2006},
  publisher={IEEE}
}

@inproceedings{niknam20233d,
  title={3d-ttp: Efficient transient temperature-aware power budgeting for 3d-stacked processor-memory systems},
  author={Niknam, Sobhan and Shen, Yixian and Pathania, Anuj and Pimentel, Andy D},
  booktitle={2023 IEEE Computer Society Annual Symposium on VLSI (ISVLSI)},
  pages={1--6},
  year={2023},
  organization={IEEE}
}

@inproceedings{pathania2018task,
  title={Task scheduling for many-cores with S-NUCA caches},
  author={Pathania, Anuj and Henkel, J{\"o}rg},
  booktitle={2018 Design, Automation \& Test in Europe Conference \& Exhibition (DATE)},
  pages={557--562},
  year={2018},
  organization={IEEE}
}

@inproceedings{gal2016dropout,
  title={Dropout as a bayesian approximation: Representing model uncertainty in deep learning},
  author={Gal, Yarin and Ghahramani, Zoubin},
  booktitle={international conference on machine learning},
  pages={1050--1059},
  year={2016},
  organization={PMLR}
}

@article{zela2019understanding,
  title={Understanding and robustifying differentiable architecture search},
  author={Zela, Arber and Elsken, Thomas and Saikia, Tonmoy and Marrakchi, Yassine and Brox, Thomas and Hutter, Frank},
  journal={arXiv preprint arXiv:1909.09656},
  year={2019}
}

@article{mohammed20233d,
  title={3D-DNaPE: Dynamic Neighbor-Aware Performance Enhancement for Thermally Constrained 3D Many-Core Systems},
  author={Mohammed, Mohammed Sultan and Al-Dhamari, Ahlam and others},
  journal={IEEE Access},
  volume={11},
  pages={131964--131978},
  year={2023},
  publisher={IEEE}
}

@article{gemma2024improving,
  title={Gemma 2: Improving Open Language Models at a Practical Size},
  author={Google Research},
  journal={arXiv preprint arXiv:2408.00118},
  year={2024}
}

@article{touvron2023llama,
  title={Llama: Open and efficient foundation language models},
  author={Touvron, Hugo and Lavril, Thibaut and Izacard, Gautier and Martinet, Xavier and Lachaux, Marie-Anne and Lacroix, Timoth{\'e}e and Rozi{\`e}re, Baptiste and Goyal, Naman and Hambro, Eric and Azhar, Faisal and others},
  journal={arXiv preprint arXiv:2302.13971},
  year={2023}
}

@article{mandal2019dynamic,
  title={Dynamic resource management of heterogeneous mobile platforms via imitation learning},
  author={Mandal, Sumit K and Bhat, Ganapati and Patil, Chetan Arvind and Doppa, Janardhan Rao and Pande, Partha Pratim and Ogras, Umit Y},
  journal={IEEE Transactions on Very Large Scale Integration (VLSI) Systems},
  volume={27},
  number={12},
  pages={2842--2854},
  year={2019},
  publisher={IEEE}
}

@article{kim2017imitation,
  title={Imitation learning for dynamic VFI control in large-scale manycore systems},
  author={Kim, Ryan Gary and Choi, Wonje and Chen, Zhuo and Doppa, Janardhan Rao and Pande, Partha Pratim and Marculescu, Diana and Marculescu, Radu},
  journal={IEEE Transactions on Very Large Scale Integration (VLSI) Systems},
  volume={25},
  number={9},
  pages={2458--2471},
  year={2017},
  publisher={IEEE}
}

@article{seeger2004gaussian,
  title={Gaussian processes for machine learning},
  author={Seeger, Matthias},
  journal={International journal of neural systems},
  volume={14},
  number={02},
  pages={69--106},
  year={2004},
  publisher={World Scientific}
}

@inproceedings{shen2022tcps,
  title={TCPS: a task and cache-aware partitioned scheduler for hard real-time multi-core systems},
  author={Shen, Yixian and Xiao, Jun and Pimentel, Andy D},
  booktitle={Proceedings of the 23rd ACM SIGPLAN/SIGBED International Conference on Languages, Compilers, and Tools for Embedded Systems},
  pages={37--49},
  year={2022}
}

@misc{intel_llama3_cpu,
  author       = {Intel},
  title        = {Meta Llama 3 Optimized CPU Inference with Hugging Face and PyTorch. },
  year         = {2024},
  url= {https://www.intel.com/content/www/us/en/developer/articles/technical/meta-llama-3-optimized-cpu-inference.html?utm_source=chatgpt.com},
}

@inproceedings{tao2024robustness,
  title={Robustness of large language models against adversarial attacks},
  author={Tao, Yiyi and Shen, Yixian and Zhang, Hang and Shen, Yanxin and Wang, Lun and Shi, Chuanqi and Du, Shaoshuai},
  booktitle={2024 4th International Conference on Artificial Intelligence, Robotics, and Communication (ICAIRC)},
  pages={182--185},
  year={2024},
  organization={IEEE}
}

@inproceedings{huang2025image2text2image,
  title={Image2text2image: A novel framework for label-free evaluation of image-to-text generation with text-to-image diffusion models},
  author={Huang, Jia-Hong and Zhu, Hongyi and Shen, Yixian and Rudinac, Stevan and Kanoulas, Evangelos},
  booktitle={International Conference on Multimedia Modeling},
  pages={413--427},
  year={2025},
  organization={Springer}
}

@article{huang2024novel,
  title={A novel evaluation framework for image2text generation},
  author={Huang, Jia-Hong and Zhu, Hongyi and Shen, Yixian and Rudinac, Stevan and Pacces, Alessio M and Kanoulas, Evangelos},
  journal={arXiv preprint arXiv:2408.01723},
  year={2024}
}

@inproceedings{huang2024optimizing,
  title={Optimizing numerical estimation and operational efficiency in the legal domain through large language models},
  author={Huang, Jia-Hong and Yang, Chao-Chun and Shen, Yixian and Pacces, Alessio M and Kanoulas, Evangelos},
  booktitle={Proceedings of the 33rd ACM International Conference on Information and Knowledge Management},
  pages={4554--4562},
  year={2024}
}

@inproceedings{aghapour2024piqi,
  title={Piqi: Partially quantized dnn inference on hmpsocs},
  author={Aghapour, Ehsan and Shen, Yixian and Sapra, Dolly and Pimentel, Andy and Pathania, Anuj},
  booktitle={Proceedings of the 29th ACM/IEEE International Symposium on Low Power Electronics and Design},
  pages={1--6},
  year={2024}
}

@inproceedings{huang2025gradient,
  title={Gradient weight-normalized low-rank projection for efficient llm training},
  author={Huang, Jia-Hong and Shen, Yixian and Zhu, Hongyi and Rudinac, Stevan and Kanoulas, Evangelos},
  booktitle={Proceedings of the AAAI Conference on Artificial Intelligence},
  volume={39},
  number={23},
  pages={24123--24131},
  year={2025}
}

@inproceedings{shen2025altgen,
  title={Altgen: Ai-driven alt text generation for enhancing epub accessibility},
  author={Shen, Yixian and Zhang, Hang and Shen, Yanxin and Wang, Lun and Shi, Chuanqi and Du, Shaoshuai and Tao, Yiyi},
  booktitle={Proceedings of the 2025 International Conference on Artificial Intelligence and Computational Intelligence},
  pages={78--83},
  year={2025}
}

@inproceedings{shen2025macp,
  title={MaCP: Minimal yet mighty adaptation via hierarchical cosine projection},
  author={Shen, Yixian and Bi, Qi and Huang, Jia-Hong and Zhu, Hongyi and Pimentel, Andy D and Pathania, Anuj},
  booktitle={Proceedings of the 63rd Annual Meeting of the Association for Computational Linguistics (Volume 1: Long Papers)},
  pages={20602--20618},
  year={2025}
}

@inproceedings{wang2025reasoning,
  title={Reasoning Beyond Points: A Visual Introspective Approach for Few-Shot 3D Segmentation},
  author={Wang, Changshuo and He, Shuting and Fang, Xiang and Hu, Zhijian and Huang, Jia-Hong and Shen, Yixian and Tiwari, Prayag},
  booktitle={The Thirty-ninth Annual Conference on Neural Information Processing Systems},
  year={2025}
}

@inproceedings{shen2025ssh,
  title={Ssh: Sparse spectrum adaptation via discrete hartley transformation},
  author={Shen, Yixian and Bi, Qi and Huang, Jia-Hong and Zhu, Hongyi and Pimentel, Andy D and Pathania, Anuj},
  booktitle={Proceedings of the 2025 Conference of the Nations of the Americas Chapter of the Association for Computational Linguistics: Human Language Technologies (Volume 1: Long Papers)},
  pages={10400--10415},
  year={2025}
}

@article{gourdoumanis2026multi,
  title={Multi-Partner Project: COIN-3D--Collaborative Innovation in 3D VLSI Reliability},
  author={Gourdoumanis, George Rafael and Oikonomou, Fotoini and Pantazi-Kypraiou, Maria and Stoikos, Pavlos and Axelou, Olympia and Tziouvaras, Athanasios and Karakonstantis, Georgios and Aladwani, Tahani and Anagnostopoulos, Christos and Shen, Yixian and others},
  journal={arXiv preprint arXiv:2601.14347},
  year={2026}
}

@inproceedings{wasala2025energy,
  title={Energy-Efficient QoS-Aware Scheduling for S-NUCA Many-Cores},
  author={Wasala, Sudam M and Wolff, Jurre and Shen, Yixian and Pathania, Anuj and Grelck, Clemens and Pimentel, Andy D},
  booktitle={2025 26th International Symposium on Quality Electronic Design (ISQED)},
  pages={1--8},
  year={2025},
  organization={IEEE}
}

@inproceedings{shenefficient,
  title={Efficient Multimodal Spatial Reasoning via Dynamic and Asymmetric Routing},
  author={Shen, Yixian and Bi, Qi and Wang, Zihan and Yang, Zhiheng and Wang, Changshuo and Zhang, Zhi and Tiwari, Prayag and Pimentel, Andy D and Pathania, Anuj},
  year={2026},
  booktitle={The Fourteenth International Conference on Learning Representations}
}

@inproceedings{zhang2025neuroada,
  title={NeuroAda: Activating Each Neuron’s Potential for Parameter-Efficient Fine-Tuning},
  author={Zhang, Zhi and Shen, Yixian and Cao, Congfeng and Shutova, Ekaterina},
  booktitle={Proceedings of the 2025 Conference on Empirical Methods in Natural Language Processing},
  pages={10960--10977},
  year={2025}
}

@inproceedings{bi2025adadcp,
  title={AdaDCP: Learning an Adapter with Discrete Cosine Prior for Clear-to-Adverse Domain Generalization},
  author={Bi, Qi and Shen, Yixian and Yi, Jingjun and Xia, Gui-Song},
  booktitle={Proceedings of the IEEE/CVF International Conference on Computer Vision},
  pages={12997--13008},
  year={2025}
}

@article{zhu2025interactive,
  title={Interactive Image Retrieval Meets Query Rewriting with Large Language and Vision Language Models},
  author={Zhu, Hongyi and Huang, Jia-Hong and Shen, Yixian and Rudinac, Stevan and Kanoulas, Evangelos},
  journal={ACM Transactions on Multimedia Computing, Communications and Applications},
  volume={21},
  number={10},
  pages={1--23},
  year={2025},
  publisher={ACM New York, NY}
}
\end{document}